\documentclass[10pt,twocolumn]{article}
\pdfoutput=1
\usepackage[a4paper,margin=0.75in]{geometry}
\usepackage[T1]{fontenc}
\usepackage[utf8]{inputenc}
\usepackage{times}
\usepackage{graphicx}
\usepackage{booktabs}
\usepackage{colortbl}
\usepackage{multirow}
\usepackage{amsmath,amssymb,bm}
\usepackage{xcolor}
\usepackage{pifont}
\usepackage{authblk}
\usepackage{cuted}
\usepackage{xspace}
\newcommand{\eg}{e.g.\xspace}

\usepackage[accsupp]{axessibility}
\usepackage[colorlinks=true,linkcolor=blue,citecolor=blue,urlcolor=blue]{hyperref}
\usepackage[capitalize,nameinlink,noabbrev]{cleveref}
\usepackage[numbers,sort&compress]{natbib}
\usepackage{stfloats}

\title{SpatialPoint: Spatial-aware Point Prediction for Embodied Localization}

\author{
Qiming Zhu$^{1,2*}$,
Zhirui Fang$^{1,2*}$,
Tianming Zhang$^{1}$,
Chuanxiu Liu$^{1}$,
Xiaoke Jiang$^{1 \dagger}$,
Lei Zhang$^{1,3}$ \\
\vspace{0.4em}
$^{1}$Visincept Research \\
$^{2}$Tsinghua University \\
$^{3}$International Digital Economy Academy (IDEA) \\
\vspace{0.2em}
{\small $^{*}$Work done during an internship at Visincept. \quad
$^{\dagger}$Corresponding author.}
}
\date{}

\begin{document}
\maketitle
\begin{abstract}
\itshape
Embodied intelligence fundamentally requires a capability to determine \textit{where} to act in 3D space. We formalize this requirement as \textbf{embodied localization} --- the problem of predicting executable 3D points conditioned on visual observations and language instructions. We instantiate embodied localization with two complementary target types: \textit{touchable points}, surface-grounded 3D points enabling direct physical interaction, and \textit{air points}, free-space 3D points specifying placement and navigation goals, directional constraints, or geometric relations. 
Embodied localization is inherently a problem of embodied 3D spatial reasoning --- yet most existing vision-language systems rely predominantly on RGB inputs, necessitating implicit geometric reconstruction that limits cross-scene generalization, despite the widespread adoption of RGB-D sensors in robotics.
To address this gap, we propose \textbf{SpatialPoint}, a spatial-aware vision-language framework with careful design that integrates structured depth into a vision-language model (VLM) and generates camera-frame 3D coordinates. We construct a 2.6M-sample RGB-D dataset covering both touchable and air points QA pairs for training and evaluation. Extensive experiments demonstrate that incorporating depth into VLMs significantly improves embodied localization performance.
We further validate SpatialPoint through real-robot deployment across three representative tasks: language-guided robotic arm grasping at specified locations, object placement to target destinations, and mobile robot navigation to goal positions. Project page: \url{https://qimingzhu-google.github.io/SpatialPoint/}.

\end{abstract}

\section{Introduction}
\label{sec:intro}
Recent advances in embodied intelligence~\cite{driess2023palm,brohan2022rt1,shridhar2022cliport} increasingly shift focus from passive visual recognition toward spatially executable perception driven by the great wave of AGI.
In many manipulation-oriented and language-conditioned robotic tasks~\cite{ahn2022can,shridhar2023perceiver}, the central requirement is to determine \textit{where} to act in 3D space. 
We argue that a broad class of embodied behaviors---ranging from grasping and placement to navigation and motion specification---can be unified under a single abstraction: predicting executable 3D points conditioned on visual observations and language instructions. 
We term this problem \textbf{embodied localization}.

\begin{figure*}[t]
  \centering
  \includegraphics[width=0.9\linewidth]{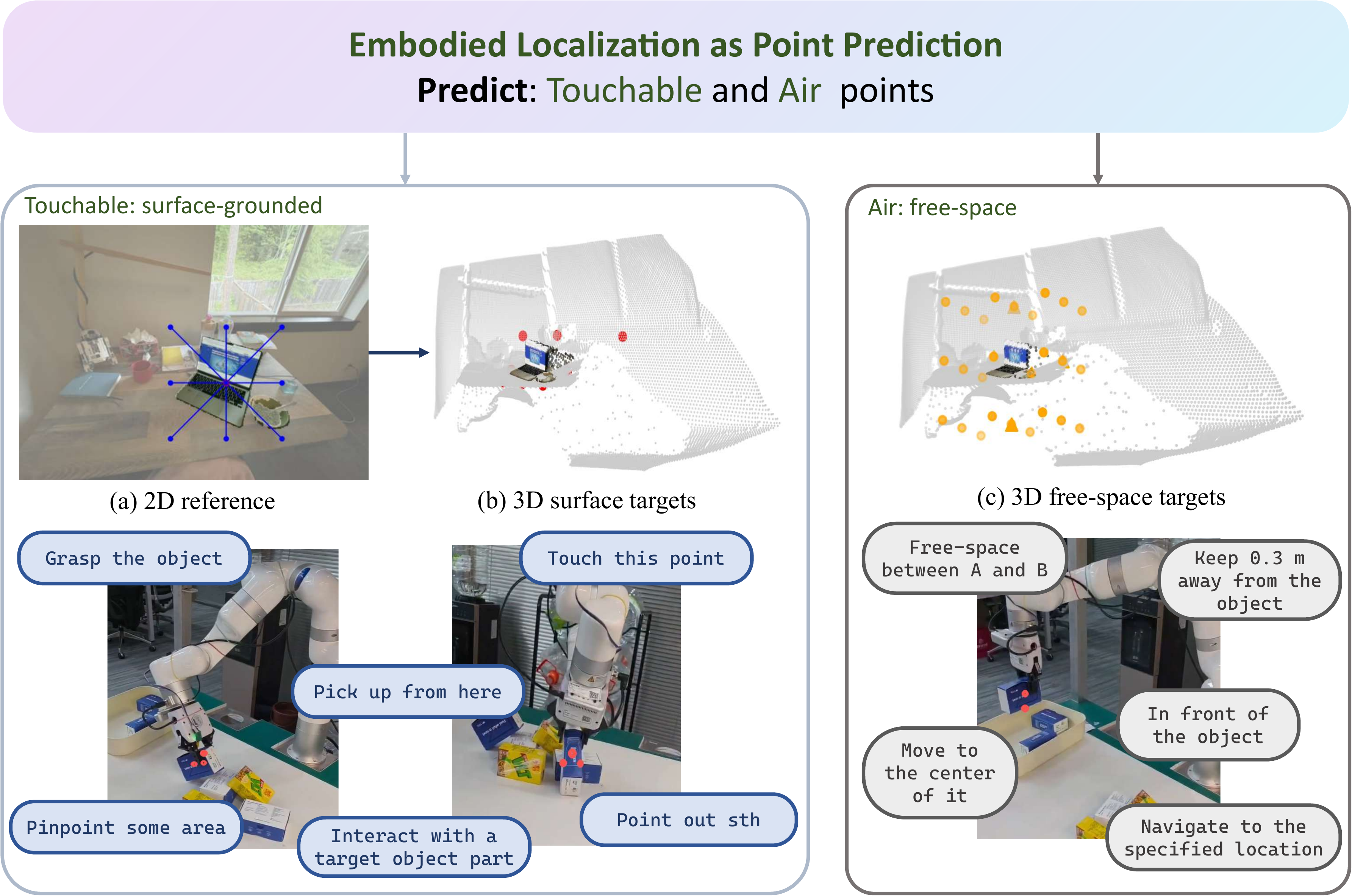}
  \caption{Embodied localization as executable 3D target prediction. We reduce embodied execution to predicting camera-frame 3D points of two complementary types: \textbf{touchable points} grounded on observed surfaces, and \textbf{air points} located in free space and specified by spatial language.}
  \label{fig:concept}
\end{figure*}

We instantiate embodied localization with two complementary target types. 
\textbf{Touchable points} are surface-grounded 3D coordinates that enable direct physical interaction, such as grasping or contact~\cite{yuan2024robopoint,qian2024thinkgrasp}. 
\textbf{Air points} are free-space 3D coordinates that specify placement and navigation goals, directional constraints, or geometric relations beyond visible surfaces~\cite{ma2024spatialpin}. 
Together, these two types constitute a minimal yet expressive representation of executable actions, unifying perception-supported and reasoning-driven targets within a single framework, as illustrated in \Cref{fig:concept}.

Embodied localization is inherently a problem of embodied 3D spatial reasoning.
However, most existing vision-language models (VLMs)~\cite{liu2024grounding,ren2024dino,bai2025qwen3} rely predominantly on RGB inputs, predicting 2D bounding boxes, masks, or other image-level outputs.
While such models can produce guesses for 3D queries, they cannot guarantee alignment with real-world metric geometry~\cite{cheng2024spatialrgpt,yang20253dmood,wang2025n3d}.
Monocular RGB cues lack explicit metric depth, often resulting in insufficient geometric fidelity for precise spatial reasoning.
Although multi-view reconstruction can recover 3D structure, it is mathematically complex and heavily dependent on accurate camera calibration and relative poses.
Consequently, learning 3D spatial structure implicitly from RGB views alone is challenging, prone to overfitting to training scenes, and often exhibits weak cross-scene generalization---despite the widespread adoption of RGB-D sensors in robotics and industrial systems that provide direct metric depth measurements.

To address this gap, we propose \textbf{SpatialPoint}, a spatial-aware vision-language framework that directly integrates structured depth into a VLM and generates camera-frame 3D coordinates via the language modeling head.
However, incorporating a new depth modality into a pre-trained vision-language model that expects only RGB and language inputs is non-trivial.
This is precisely why most existing approaches avoid using depth at the input stage, instead treating it as an auxiliary cue for feature augmentation or post-hoc refinement (see \Cref{related_depth}).
To tackle this modality integration challenge, SpatialPoint takes RGB and structured depth as parallel inputs: depth is first encoded into a 3-channel representation and processed by a dedicated depth backbone, producing depth tokens that are fused with visual and language features within the multimodal backbone for depth-aware reasoning.
To fully activate the depth modality, we further adopt a two-stage training strategy that progressively aligns the depth encoder with the pre-trained VLM.
Together, these careful design choices are what enable structured depth to contribute meaningfully to embodied 3D spatial reasoning, rather than introducing noise or degrading the pre-trained capabilities.

To enable large-scale training and evaluation, we construct \textbf{SpatialPoint-Data}, a 2.6M-sample RGB-D dataset covering both touchable and air points question-answer pairs. 
Touchable targets are obtained by lifting 2D interaction annotations into camera-frame 3D coordinates using depth and camera geometry. 
Air targets are automatically synthesized to cover direction-specified and relation-centric queries with optional metric constraints.
We further establish \textbf{SpatialPoint-Bench} for unified evaluation over both target types. 
Extensive experiments demonstrate that incorporating structured depth into VLMs consistently improves embodied localization performance over RGB-only baselines and alternative geometry-input designs, establishing depth tokens as a critical geometric inductive bias for scalable, spatial-aware embodied reasoning.

Finally, we validate SpatialPoint through real-robot deployment across three representative tasks: language-guided robotic arm grasping at specified locations (\textit{touchable points}), object placement to target destinations (\textit{air points}), and mobile robot navigation to goal positions (\textit{air points}).
The consistently strong real-world performance demonstrates the practical effectiveness and generalizability of our approach. Our contributions are threefold:
\begin{enumerate}
    \item We introduce embodied localization as a unified formulation of executable 3D target prediction and instantiate it with two complementary types: touchable points and air points.
    \item We construct SpatialPoint-Data (2.6M samples) and SpatialPoint-Bench, enabling large-scale training and standardized evaluation of depth-aware embodied localization.
    \item We propose SpatialPoint, an RGB-D extension of a VLM with an explicit depth-token stream that directly generates camera-frame 3D coordinates, achieving consistent gains in spatial reasoning and cross-scene generalization, validated by both offline benchmarks and real-robot experiments.
\end{enumerate}

\section{Related Work}
\subsection{Spatial Understanding in Vision-Language Models}

Recent vision-language models (VLMs) have progressed from 2D grounding toward spatial understanding in 3D environments.
A common direction is to explicitly inject 3D representations into multimodal frameworks~\cite{yang2024llm,chen2024grounded}, \eg, aligning point-cloud features with large language models for 3D grounding, or improving grounding and relational reasoning via fine-grained reward modeling~\cite{yan2024vigor}.
Another line preserves strong 2D backbones while enabling 3D reasoning through multi-view inputs or coordinate prompting~\cite{guo2023viewrefer,xu2024vlm,liu20253daxisprompt}.

Hybrid approaches further incorporate geometric cues such as depth, lifting, or spatial programs.
SpatialRGPT~\cite{cheng2024spatialrgpt} introduces a depth plugin that leverages monocular (relative) depth to improve direction and distance reasoning; ZSVG3D~\cite{yuan2024visual} formulates spatial reasoning as visual programs; and several works lift 2D detections into 3D space for object-level localization~\cite{li2025seeground,yang20253dmood}.
Recent methods also explore more expressive 3D representations—such as 3D Gaussians and structured scene abstractions~\cite{liu2025reasongrounder,zemskova20253dgraphllm,thai2025splattalk,qi2025gpt4scene}—to support relational reasoning.

Despite these advances, most approaches remain object-centric, focusing on 3D boxes, scene graphs, or coarse abstractions, rather than producing fine-grained \emph{executable} spatial targets for embodied interaction.

\subsection{Vision-Language Interfaces for Manipulation Execution}

Embodied tasks ultimately require actionable targets in metric 3D space.
One line of work~\cite{yuan2024robopoint} predicts language-conditioned interaction points or affordances, emphasizing contact-level grounding for manipulation.
Another line represents actions in structured 3D spaces to better couple perception and control, such as voxelized action prediction from RGB-D reconstructions~\cite{shridhar2023perceiver} or sequence-level control generation in continuous action spaces~\cite{chi2025diffusion,zhao2023learning}.

More recently, vision-language-action (VLA) models scale end-to-end policy learning for multi-task generalization.
Systems such as RT-1/RT-2~\cite{brohan2022rt1,zitkovich2023rt2}, PaLM-E~\cite{driess2023palm}, and SayCan~\cite{ahn2022can} integrate language, vision, and control at different levels of abstraction, while open generalist policy backbones---including OpenVLA and Octo~\cite{kim2024openvla,team2024octo}---are trained on large embodied datasets.
Complementary to policy learning, explicit 3D world modeling offers geometry-grounded action representations; for example, PointWorld~\cite{huang2026pointworld} predicts 3D point flows in a shared metric space for manipulation.

On the dataset side, RoboAfford~\cite{tang2025roboafford} scales affordance supervision for manipulation, emphasizing object parts and surface contact regions.
At a broader scope, RoboBrain 2.5~\cite{tan2026RoboBrain} aggregates multi-source embodied supervision to support generalist models, including depth-aware coordinate-related signals.

However, existing approaches typically focus either on surface-supported interaction points or on end-to-end action generation, and rarely provide a unified view that covers both \emph{surface-attached} targets and reasoning-required goals in surrounding free space under a consistent vision-conditioned interface.

\subsection{Depth Cues for Spatial Reasoning in VLMs and VLAs}
\label{related_depth}
In practical robotics, depth is often readily available via low-cost RGB-D sensors~\cite{chen2018calibrate,rustler2025empirical} (or can be approximated by monocular depth estimation~\cite{lin2025da3}), making metric geometry an economical signal rather than an extra burden.
Accordingly, depth cues provide a direct way to inject metric structure into vision-language models and embodied policies, improving direction and distance reasoning beyond what RGB-only cues can reliably support.

For VLMs, SpatialBot~\cite{cai2025spatialbot} explicitly consumes RGB and depth images to enhance depth-aware spatial understanding,
while SpatialRGPT~\cite{cheng2024spatialrgpt} introduces a flexible depth plugin that incorporates (typically estimated) depth maps as auxiliary inputs to existing VLM visual encoders and benchmarks.
For VLAs, DepthVLA~\cite{yuan2025depthvla} explicitly incorporates depth prediction into policy architectures to improve manipulation robustness, while PointVLA and GeoVLA~\cite{li2026pointvla,sun2025geovla} inject point-cloud or depth-derived geometric features to enhance spatial generalization.
Beyond adding depth as an auxiliary cue, 3D-VLA~\cite{zhen20243dvla} links 3D perception, reasoning, and action through a 3D-centric foundation model, and LLaVA-3D~\cite{zhu2024llava} adapts large multimodal models toward 3D outputs via 3D-augmented visual patches.

These efforts collectively support the view that explicit depth cues are valuable for metric spatial reasoning.
In contrast to end-to-end action generation, our depth-aware vision-language modeling directly produces \emph{camera-frame 3D target points} and evaluates both \emph{touchable} and \emph{air} targets with complementary metrics.

\section{Methodology}
\label{sec:method}

We formulate embodied localization as \emph{depth-aware, language-conditioned 3D point prediction}. Given an RGB image, a depth map, and an instruction, the model generates a list of 3D target points.
This section presents the construction of SpatialPoint-Data, which serves as the fuel for introducing a new depth modality into vision-language models; followed by the network architecture and training strategy of SpatialPoint; and finally the evaluation protocol, including the proposed metrics and the benchmark dataset SpatialPoint-Bench. \cref{tab:dataset} summarizes the composition of SpatialPoint-Data across both touchable-point and air-point supervision.

\subsection{SpatialPoint-Data and Data Engine}
\label{sec:targets}

\begin{table}[t]
\centering
\caption{Composition of SpatialPoint-Data and SpatialPoint-Bench. Ratios are computed within each category.}
\label{tab:dataset}
\small
\setlength{\tabcolsep}{4.4pt}
\begin{tabular}{llrr}
\hline
\textbf{Category} & \textbf{Subtype} & \textbf{\#QA} & \textbf{Ratio (\%)} \\
\hline
\multirow{5}{*}{Touchable-Data} 
& Object detection & 513{,}395 & 26.99 \\
& Object pointing & 161{,}681 & 8.50 \\
& Object affordance & 560{,}836 & 29.48 \\
& Object reference & 346{,}617 & 18.22 \\
& Region reference & 319{,}961 & 16.82 \\
\hline
\multirow{3}{*}{Touchable-Bench}
& Object affordance & 124 & 36.69 \\
& Spatial affordance & 100 & 29.59 \\
& Object reference & 114 & 33.73 \\
\hline
\multirow{5}{*}{Air-Data}
& direction only & 252{,}267 & 35.18 \\
& direction (offset) & 241{,}373 & 33.66 \\
& body-length & 138{,}615 & 19.33 \\
& between & 26{,}981 & 3.76 \\
& between (offset) & 57{,}890 & 8.07 \\
\hline
\multirow{5}{*}{Air-Bench}
& direction only & 927 & 37.91 \\
& direction (offset) & 855 & 34.97 \\
& body-length & 422 & 17.26 \\
& between & 72 & 2.94 \\
& between (offset) & 169 & 6.91 \\
\hline
\end{tabular}
\end{table}

\paragraph{Data source.}

\begin{figure*}[t]
  \centering
  \includegraphics[width=.9\linewidth]{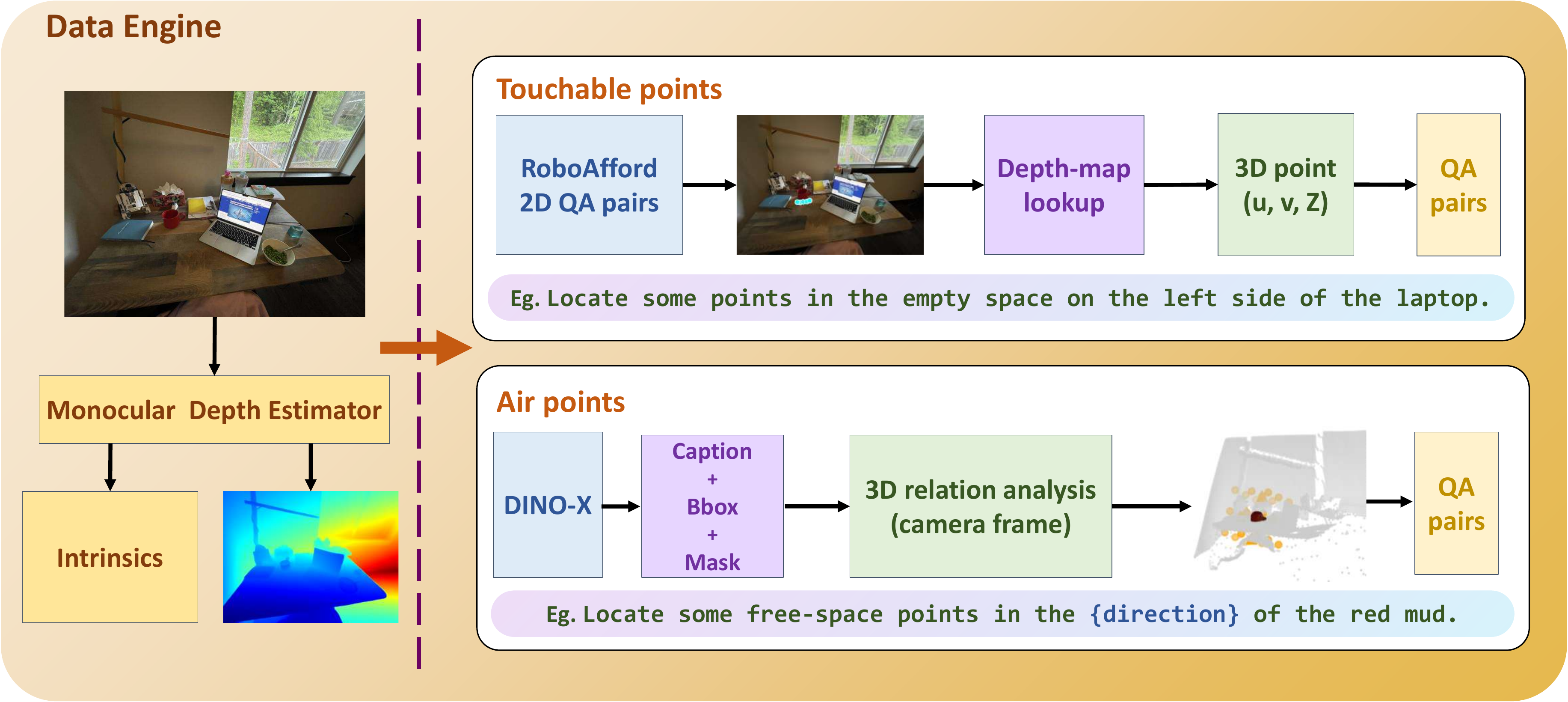}
  \caption{Our data engine. Touchable points are converted from RoboAfford~\cite{tang2025roboafford} 2D annotations using monocular depth estimation~\cite{lin2025da3}. Air points are generated from objects' 3D relations, which are computed by lifting DINO-X~\cite{ren2024dino} detections (caption/bbox/mask) into the camera frame using the estimated depth map and intrinsics, and then applying geometric computations.}
  \label{fig:dataengine}
\end{figure*}

Since introducing a new modality as input to VLM is non-trivial and massive data is the best fuel, we build SpatialPoint-Data on RoboAfford-Data~\cite{tang2025roboafford}, due to its large scale, diverse manipulation-related task types, and publicly available RGB imagery.
\cref{fig:dataengine} provides an overview of our data engine, as explained in the following text.

Each target is encoded as a triplet $(u,v,Z)$, where $(u,v)$ are pixel coordinates in the image frame and $Z$ is pixel depth in millimeters.
In our implementation, $u$ and $v$ are integers in $[0,1000)$, and $Z$ is an integer-valued depth.
Targets are serialized as plain text and learned via standard token prediction, without additional coordinate binning.
We then use off-the-shelf monocular depth estimation model~\cite{lin2025da3} to estimate depth for each RGB image.

\paragraph{Touchable points via monocular depth estimation.}
\label{sec:touchable_targets}
RoboAfford provides 2D QA pairs with surface-attached supervision, where each query is associated with valid 2D points on the image.
Therefore, we first obtain the 2D target location $(u,v)$ from the 2D annotation, look up the corresponding depth value $Z$ from the estimated depth map at $(u,v)$, and form the unified 3D target $(u,v,Z)$.
By applying this lift-from-2D procedure to all touchable points samples, we construct \textbf{1.9M} touchable points QA pairs.


\paragraph{Air points via 3D relation analysis.}
\label{sec:air_targets}
Air points are not anchored to annotated surface contacts. Instead, we generate QA pairs from RGB images using estimated depth and geometry constraints derived from object-centric 3D context.
Concretely, we first parse each image with DINO-X~\cite{ren2024dino} to obtain objects' captions, bounding boxes, and masks.
Combining these outputs with the estimated depth map and intrinsics, we lift the scene into the camera frame to obtain object-centric 3D occupancy cues.
Based on this geometry, we can directly derive objects' 3D relations via geometric computations.
Using the resulting 3D relations, we construct \textbf{0.72M} QA pairs over \textbf{26} discrete camera-frame directions (six canonical axes and their compositions) and organize them into three groups:
(i) \emph{direction} queries that ask for a point in a specified direction (optionally with a metric offset),
(ii) \emph{between-object} queries that place a point between two referenced objects, optionally with distance constraints (\eg, \emph{near} one object or with a metric offset), and
(iii) \emph{body-length} queries that mirror the direction form but express distance using body-length units, requiring the model to reason about the target object's physical extent.

\begin{figure}[t]
  \centering
  \includegraphics[width=1.0 \linewidth]{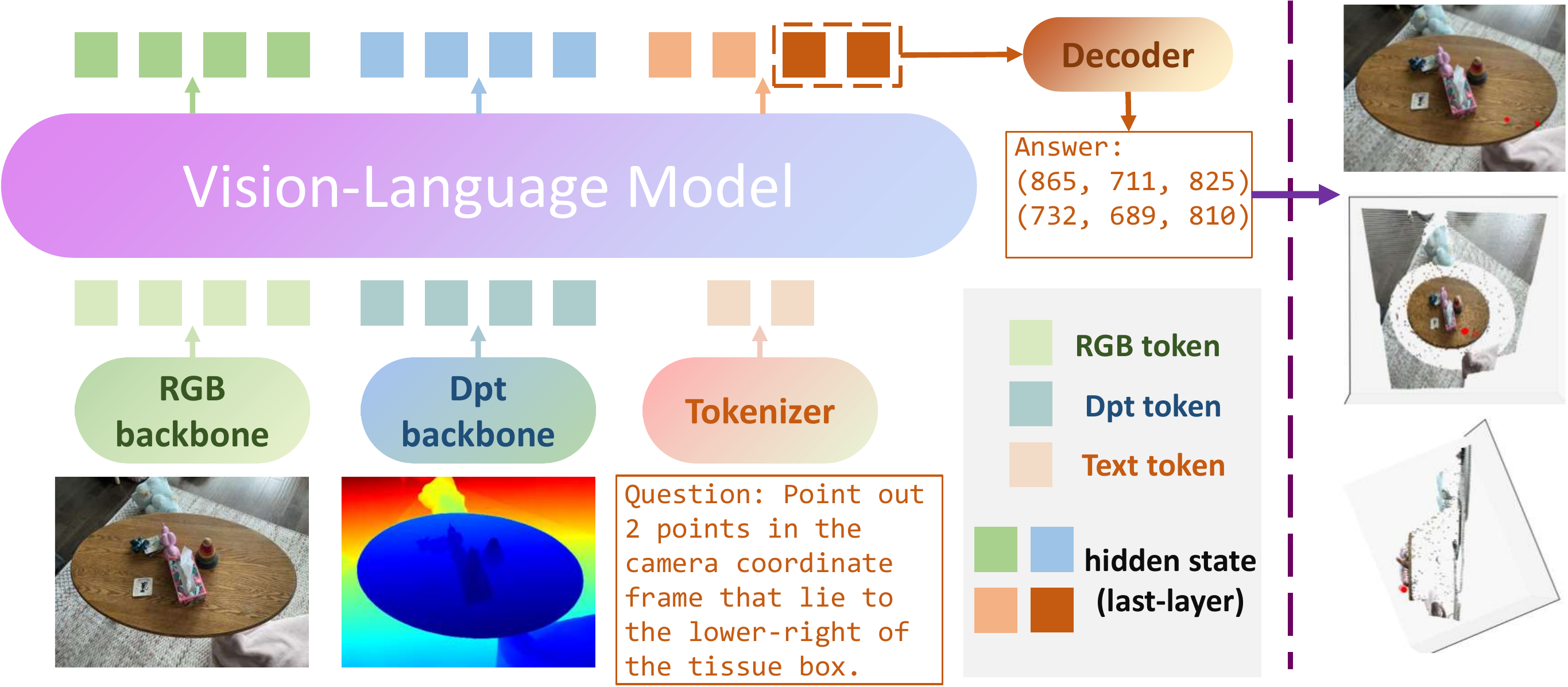}
\caption{Model overview. We add a dedicated depth encoder by duplicating the original visual backbone and feeding it a three-channel depth map to obtain depth tokens, wrapped by \texttt{<dpt\_start>} and \texttt{<dpt\_end>}. RGB/depth/text tokens form one causal sequence, and the LM head decodes structured \((u,v,Z)\) point lists.}
  \label{fig:arch}
\end{figure}

\begin{table*}[t]
\centering
\caption{Touchable points results on RoboAfford-Eval. We report overall accuracy and per-category accuracy for object affordance prediction (OAP), object affordance recognition (OAR), and spatial affordance localization (SAL). For methods that output $(u,v,Z)$ targets, we additionally report depth MAE (mm) against the monocular depth reference used in our lifted-3D evaluation, with an inside/outside breakdown; otherwise depth metrics are marked as ``$-$''.}
\label{tab:surface_results}
\tiny
\setlength{\tabcolsep}{3.6pt}
\renewcommand{\arraystretch}{1.08}

\resizebox{\textwidth}{!}{%
\begin{tabular}{lccccccc}
\toprule
\textbf{Method} 
& \textbf{Overall$\uparrow$}
& \textbf{OAP$\uparrow$}
& \textbf{OAR$\uparrow$}
& \textbf{SAL$\uparrow$}
& \textbf{MAE$_Z$(in)$\downarrow$}
& \textbf{MAE$_Z$(out)$\downarrow$}
& \textbf{MAE$_Z$(all)$\downarrow$} \\
\midrule
RoboPoint\cite{yuan2024robopoint}  & 0.447 & 0.350 & 0.557 & 0.442 & $-$ & $-$ & $-$ \\
RoboAfford-Qwen\cite{tang2025roboafford}  & 0.593 & 0.453 & 0.661 & 0.579 & $-$ & $-$ & $-$ \\
RoboAfford-Qwen$++$\cite{hao2025roboaffordplus}  & 0.634 & 0.631 & 0.705 & 0.558 & $-$ & $-$ & $-$ \\
RoboBrain 2.5\cite{tan2026RoboBrain}              & 0.741 & 0.673 & 0.876 & 0.671 & 205.1 & 255.9 & 217.9 \\
Qwen2.5-VL-7B\cite{Bai2025Qwen25VLTR}             & 0.161 & 0.083 & 0.195 & 0.219 & $-$ & $-$ & $-$ \\
Qwen3-VL-Inst-2B\cite{bai2025qwen3}          & 0.319 & 0.307 & 0.445 & 0.190 & $-$ & $-$ & $-$ \\
Qwen3-VL-Inst-4B\cite{bai2025qwen3}          & 0.503 & 0.573 & 0.540 & 0.375 & 567.1 & 577.9 & 574.8 \\
\rowcolor{gray!20} 
Ours           & \textbf{0.790} & \textbf{0.785} & \textbf{0.885} & \textbf{0.689} & \textbf{9.3}   & \textbf{54.4}  & \textbf{17.2} \\
\bottomrule
\end{tabular}
}
\end{table*}

\subsection{Model Architecture}
Given an RGB image $I$, a depth map $D$, and an instruction $T$, the input sequence is composed of RGB visual tokens, depth visual tokens, and text tokens. RGB and depth tokens are enclosed by dedicated delimiters. Conditioned on this RGB-D prefix, the model predicts a variable-length list of camera-centric targets in the structured form $[(u,v,Z), \ldots]$.

\begin{table}[t]
\centering
\caption{Overall air points evaluation on SpatialPoint-Bench (2445 queries). }
\label{tab:free_overall}
\small
\setlength{\tabcolsep}{4.6pt}
\renewcommand{\arraystretch}{1.05}
\resizebox{1.0\columnwidth}{!}{%
\begin{tabular}{lccc}
\toprule
\textbf{Method} & \textbf{DirPt} & \textbf{MetPt@5cm} & \textbf{MeanErr (cm)} \\
\midrule
RoboBrain 2.5\cite{tan2026RoboBrain}             & 0.0804 & 0.0637 & 30.3412 \\
Qwen3-VL-Inst-4B\cite{bai2025qwen3}         & 0.0532 & 0.0896 & 54.7086 \\
Ours (epoch 1)           & 0.4886 & 0.2587 & 8.5008 \\
Ours (epoch 2)           & \textbf{0.5088} & \underline{0.2907} & \underline{7.3034} \\
\rowcolor{gray!20} 
Ours (epoch 3)           & \underline{0.5071} & \textbf{0.3347} & \textbf{6.8084} \\
\bottomrule
\end{tabular}
}
\end{table}

\paragraph{\textbf{Depth map encoding.}}
Inspired by SpatialBot~\cite{cai2025spatialbot}, we encode the single-channel depth map into a three-channel uint8 depth image, making depth compatible with the vision tokenizer while preserving geometric structure.

\paragraph{\textbf{Dual backbones for RGB and depth.}}
Since the depth map constitutes a new visual modality, we introduce a dedicated depth backbone while keeping the rest of the model unchanged.
To maximize modality alignment, we directly duplicate the RGB visual backbone and allocate it exclusively to depth, using the same architecture but separate parameters.
Both backbones operate on the same patch grid, yielding token sequences that are naturally aligned for subsequent joint reasoning.
To make the depth backbone effective without destabilizing the pretrained components, we adopt a two-stage training strategy.
In Stage~1, we freeze all modules except the depth backbone and train only this backbone with a $10\times$ larger learning rate.
In Stage~2, we unfreeze the full model and perform joint finetuning with the standard learning rate.

\paragraph{\textbf{Causal multimodal fusion and structured decoding.}}
We follow the standard causal VLM interface that organizes non-text modalities as explicit segments in a single token sequence. 
In typical VLMs, image patch tokens are inserted into the prompt and bracketed by dedicated boundary markers to make the visual prefix explicit without changing the underlying attention pattern.
Following the same design principle, we introduce a symmetric pair of special tokens, \texttt{<dpt\_start>} and \texttt{<dpt\_end>}, to delimit the depth segment, within which we place the depth-backbone patch tokens.
These non-overlapping marker spans keep RGB and depth explicit in the multimodal prefix while preserving the original causal architecture.
Conditioned on the RGB--depth prefix and the instruction tokens, the model autoregressively generates a bracketed list of 3D targets with a fixed syntax (\eg, \texttt{[(123, 456, 789), ...]}).
We then deterministically parse the generated string into $(u,v,Z)$ tuples for evaluation and downstream geometric computation.

\begin{table*}[t]
\centering
\caption{Point-level micro results by category. MetPt is computed on direction-correct points in metric-offset queries (denominator: dir-passed points in has\_metric samples).}
\label{tab:free_bycat_micro}
\tiny
\setlength{\tabcolsep}{3.6pt}
\renewcommand{\arraystretch}{1.1}

\resizebox{\textwidth}{!}{%
\begin{tabular}{lccc ccc ccc}
\toprule
& \multicolumn{3}{c}{\textbf{Direction$\uparrow$}} 
& \multicolumn{3}{c}{\textbf{Between$\uparrow$}} 
& \multicolumn{3}{c}{\textbf{Body-length$\uparrow$}} \\
\cmidrule(lr){2-4} \cmidrule(lr){5-7} \cmidrule(lr){8-10}
\textbf{Method} 
& \textbf{DirPt} & \textbf{MetPt} & \textbf{FullPt}
& \textbf{DirPt} & \textbf{MetPt} & \textbf{FullPt}
& \textbf{DirPt} & \textbf{MetPt} & \textbf{FullPt} \\
\midrule
Qwen3-VL-Inst-4B\cite{bai2025qwen3} & 0.0573 & 0.1111 & 0.0050 & 0.0108 & 0.2222 & 0.0024 & 0.0627 & 0.0175 & 0.0011 \\
RoboBrain 2.5\cite{tan2026RoboBrain}        & 0.0817 & 0.0827 & 0.0070 & 0.0543 & 0.0333 & 0.0018 & 0.0904 & 0.0455 & 0.0041 \\
Ours (epoch 1)      & 0.4856 & \underline{0.3096} & \underline{0.1571} & 0.3825 & 0.1833 & 0.0701 & \textbf{0.5637} & 0.1959 & 0.1104 \\
Ours (epoch 2)      & \textbf{0.5264} & 0.2864 & 0.1523 & \underline{0.4031} & \textbf{0.2711} & \textbf{0.1093} & 0.4964 & \underline{0.3093} &\underline{ 0.1535} \\
\rowcolor{gray!20} 
Ours (epoch 3)      & \underline{0.5161} & \textbf{0.3764} & \textbf{0.1867} & \textbf{0.4371} & \underline{0.2123} & \underline{0.0928} & \underline{0.5097} & \textbf{0.3143} & \textbf{0.1602} \\
\bottomrule
\end{tabular}
}
\end{table*}

\subsection{SpatialPoint-Bench}
\label{sec:bench}

\paragraph{\textbf{Evaluation data.}}
We evaluate on \textbf{SpatialPoint-Bench}, whose images are sourced from the real-scene split of RoboAfford-Eval~\cite{tang2025roboafford}.
For each image, we predict a dense depth map using the same monocular depth estimator as in training, and use it to express both predictions and targets in the unified $(u,v,Z)$ format.

For touchable points evaluation, RoboAfford-Eval~\cite{tang2025roboafford} provides a ground-truth valid 2D region mask for each query.
By pairing this mask with the estimated depth at the corresponding locations, we obtain surface-attached $(u,v,Z)$ targets and a well-defined touchable points benchmark.

For air points evaluation, we generate air points QA pairs by following the same geometry-constrained synthesis procedure used for training (\cref{sec:air_targets}).

\paragraph{\textbf{Touchable points: 2D accuracy from region masks.}}
Each model response yields a set of predicted points $\{(\hat{u}_i,\hat{v}_i,\hat{Z}_i)\}_{i=1}^{N}$.
Given the valid region mask $M$, per-query 2D accuracy is defined as the fraction of predicted $(\hat{u}_i,\hat{v}_i)$ that fall inside the valid region:
\begin{equation}
\mathrm{Acc}_{2D} = \frac{1}{N}\sum_{i=1}^{N}\mathbb{I}\big[M(\hat{u}_i,\hat{v}_i)=1\big].
\label{eq:acc2d}
\end{equation}

\paragraph{\textbf{Touchable points: depth error (MAE) against depth reference.}}
We evaluate depth by comparing $\hat{Z}_i$ against the depth reference value $Z^{\mathrm{ref}}(\hat{u}_i,\hat{v}_i)$ at the predicted location, and report mean absolute error (mm):
\begin{equation}
\mathrm{MAE}_{Z} = \frac{1}{N}\sum_{i=1}^{N}\left|\hat{Z}_i - Z^{\mathrm{ref}}(\hat{u}_i,\hat{v}_i)\right|.
\label{eq:maez}
\end{equation}
We additionally report $\mathrm{MAE}_{Z}$ on predictions \emph{inside} the mask, \emph{outside} the mask, and \emph{overall}.

\paragraph{\textbf{Air points: direction correctness.}}
We evaluate whether a prediction satisfies the language-specified geometric relation in the camera frame.
Let $\mathbf{c}$ denote the anchor (proxy-box center) of the referenced object, and $\mathbf{p}$ be the predicted 3D point converted by back-projecting pixel $(u,v,Z)$ in the image to 3D space.
For \emph{direction} queries with a direction unit vector $\mathbf{d}$, we mark direction as correct if the prediction lies within a conic sector, i.e.,  
$\angle(\mathbf{p}-\mathbf{c},\, \mathbf{d}) \leq \alpha$, where $\alpha$ is a fixed angular tolerance.


For \emph{between-object} queries, we define a cylindrical corridor along the segment connecting the two object anchors $\mathbf{c}_A$ and $\mathbf{c}_B$.
A prediction is marked correct if it lies within a radius-$\rho$ cylinder around the segment and projects onto the segment (within endpoints), capturing the "in-between" constraint.

\paragraph{\textbf{Occupation exclusion.}}
To enforce "not occupied by other objects", we reject predictions that fall inside any other object's inflated proxy box (constructed from lifted mask point clouds under estimated depth).

\paragraph{\textbf{Air points: distance bias error (conditional).}}
For queries that include an explicit distance constraint, we evaluate distance only when the direction or relationship constraint is satisfied.
Let $r=\|\mathbf{p}-\mathbf{c}\|_2$ be the predicted displacement magnitude (with anchor $\mathbf{c}$ chosen per query type) and $r^\ast$ be the required magnitude.
We report distance bias as
\begin{equation}
\mathrm{Bias}_{dist} = |r - r^\ast|.
\label{eq:dist_bias}
\end{equation}

\paragraph{\textbf{Body-length definition.}}
To reduce sensitivity to mask noise, we define the body-length scale from the coarse 3D proxy box:
the body length is set to half of the proxy box diagonal length.
Accordingly, $r^\ast$ in \cref{eq:dist_bias} is specified in this normalized scale for body-length queries, and in metric units for metric-offset queries.

\section{Experiments}
\label{sec:exp}
In this section, we begin with training details and evaluation metrics, then present the main experimental results, followed by ablation studies and qualitative visualizations. Beyond offline evaluation, we further conduct real-robot experiments to validate practical effectiveness; the experimental setup, results, and video demonstrations are available in the supplementary material.

\subsection{Experimental Setup}
\label{sec:exp_setup}
\paragraph{\textbf{Dataset.}}
Our network is trained with SpatialPoint-Data which includes approximately \textbf{1.9M} touchable points samples and \textbf{0.72M} air points samples. 
Our offline evaluation is conducted on SpatialPoint-Bench.





\paragraph{\textbf{Model training.}}
We build upon Qwen3-VL-Inst-4B~\cite{bai2025qwen3} and introduce an explicit depth-token stream while keeping the original causal multimodal transformer unchanged (see \cref{fig:arch}). We adopt a two-stage optimization scheme for the newly introduced depth branch.
We first warm up the depth branch by training only the duplicated depth vision backbone while freezing the VLM, the RGB vision backbone, and the multimodal transformer; during this stage, we use a learning rate that is \textbf{10$\times$} the base learning rate.
We then unfreeze all components and jointly fine-tune the entire model with AdamW and cosine learning-rate decay to perform instruction-tuned point generation under mixed surface and air points supervision.
Unless otherwise specified, we train for 1 epoch.
All experiments are run on 8 GPUs with per-GPU batch size 4 and gradient accumulation 2.
During inference, we decode the coordinate list from the LM head with \texttt{top-$p$} sampling ($p=0.9$), \texttt{top-$k$} sampling ($k=50$), and temperature $0.1$.


To study longer adaptation on air points targets, we additionally continue fine-tuning on the air points subset for 1/2/3 epochs under the same architecture and recipe, and report the corresponding checkpoints in \cref{tab:free_overall,tab:free_bycat_micro}.

\begin{figure*}[!t]
  \centering
  \includegraphics[width=0.9\linewidth]{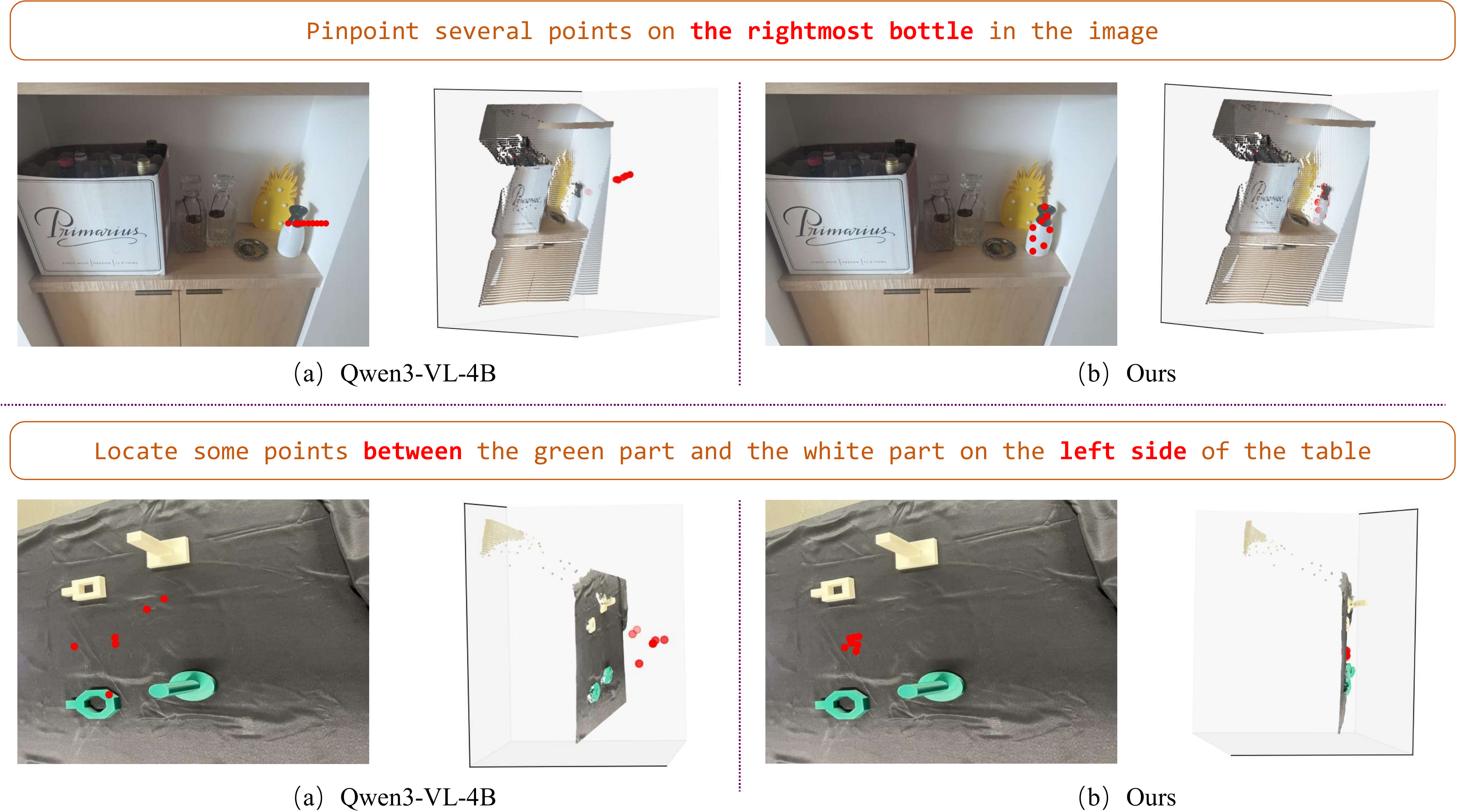}
  \caption{Surface-target qualitative comparison on RoboAfford-Eval(touchable-point).}
  \label{fig:qual_surface}
\end{figure*}

\begin{figure*}[b]
  \centering
  \includegraphics[width=0.9\linewidth]{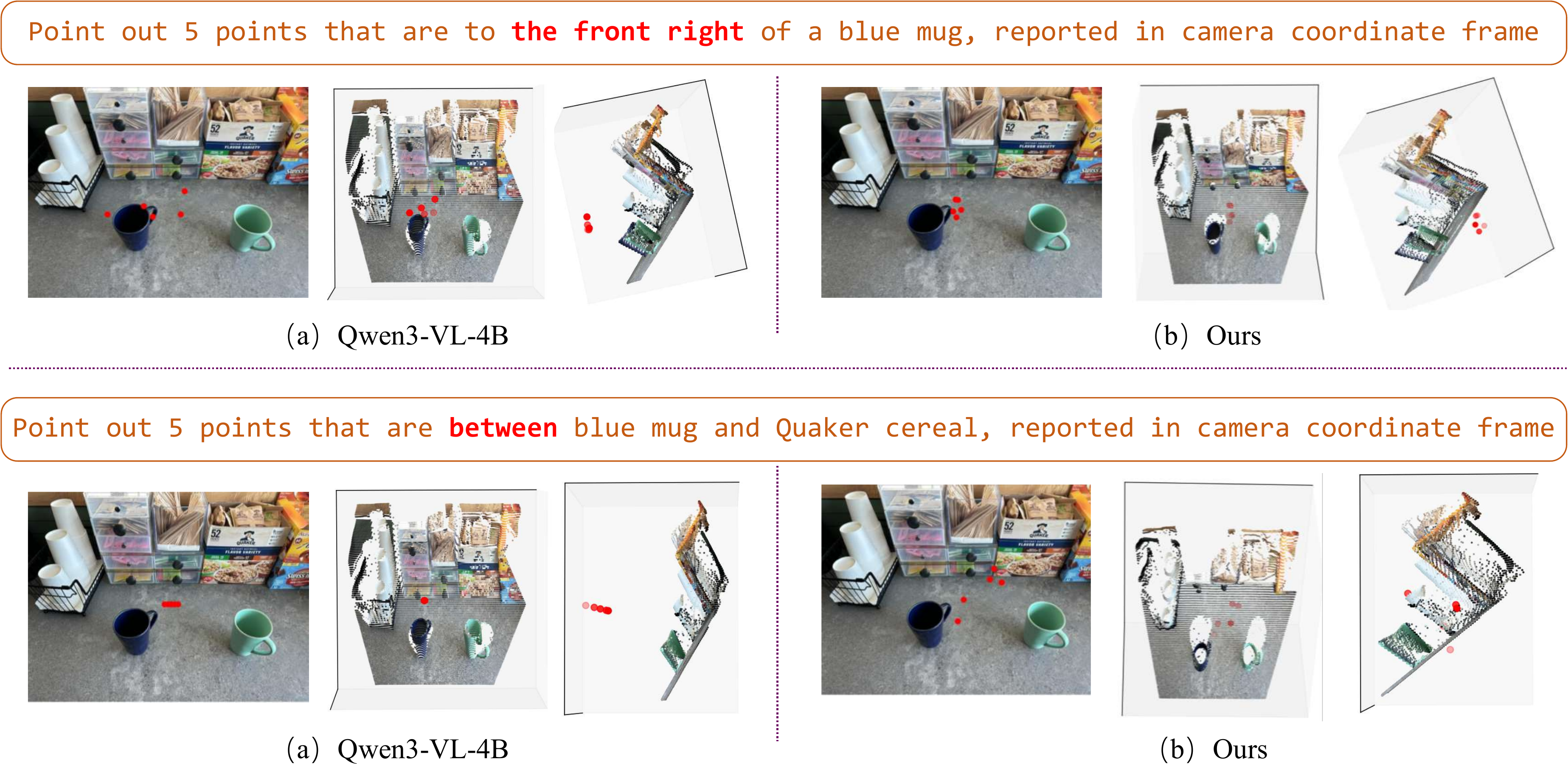}
  \caption{Free-space qualitative comparison on our benchmark. Under the same queries, our model better satisfies air points relation constraints, compared to Qwen3-VL, demonstrating more reliable air points target prediction.}
  \label{fig:qual_free}
\end{figure*}

\begin{table*}[t]
\centering
\caption{Touchable points ablations on RoboAfford-Eval. We vary the extra 3D input modality (depth map vs. point cloud), whether using a dual-branch backbone, whether depth is encoded in the SpatialBot-style (Depth$\rightarrow$3ch) when applicable, and whether special depth tokens \texttt{<dpt\_start>}/\texttt{<dpt\_end>} are used. Depth MAE is reported only when valid depth predictions are available under our lifted-3D evaluation; otherwise it is marked as ``--''.}
\label{tab:ablation_surface}
\tiny
\setlength{\tabcolsep}{3.2pt}
\renewcommand{\arraystretch}{1.08}

\resizebox{\textwidth}{!}{%
\begin{tabular}{lcccccc}
\toprule
\textbf{Model}
& \textbf{Extra input}
& \textbf{Dual-branch}
& \textbf{SpatialBot enc.}
& \textbf{Special tokens}
& \textbf{Overall Acc$\uparrow$}
& \textbf{Depth MAE (mm)$\downarrow$} \\
\midrule
\textbf{Ours e1}   & Depth map  & \checkmark & \checkmark  & \checkmark & \textbf{0.790} & \textbf{17.2} \\
\midrule
T1 & --  & \texttimes  & \texttimes  & \texttimes & 0.676 & 227.1 \\
T2 & --  & \texttimes  & \texttimes  & \checkmark & 0.698 & 191.9 \\
T3 & Depth map  & \texttimes & \checkmark  & \texttimes & 0.708 & 49.2 \\
T4      & Depth map  & \texttimes & \checkmark  & \checkmark  & 0.707 & 22.1 \\
T5      & Depth map  & \checkmark & \texttimes  & \checkmark  & 0.736 & 38.1 \\
T6     & Point cloud& \checkmark & --          & \checkmark & 0.443 & 23.3 \\
\bottomrule
\end{tabular}
}
\end{table*}

\begin{table*}[t]
\centering
\caption{Air points ablations on SpatialPoint-Bench (overall). We fix the dual-branch backbone and special depth tokens (used only when an extra input is provided), and only vary the input representation.}
\label{tab:ablation_free}
\tiny
\setlength{\tabcolsep}{4.2pt}
\renewcommand{\arraystretch}{1.08}

\resizebox{\textwidth}{!}{%
\begin{tabular}{lllcccc}
\toprule
\textbf{Variant} & \textbf{Epoch} & \textbf{Input} &
\textbf{DirPt$\uparrow$} & \textbf{MetPt@5cm$\uparrow$} &
\textbf{FullPt$\uparrow$} & \textbf{MeanErr (cm)$\downarrow$} \\
\midrule
Ours1 & 1 & SpatialBot-encoded depth & 0.4886 & 0.2587 & 0.1300 & 8.5008  \\
Ours2 & 2 & SpatialBot-encoded depth & \textbf{0.5088} & 0.2907 & 0.1456 & 7.3034  \\
Ours3 & 3 & SpatialBot-encoded depth & 0.5071 & 0.3347 & 0.1641 & 6.8084  \\
\midrule
A1 & 1 & rgb-only  & 0.2007 & 0.0810 & 0.0181 & 18.6502 \\
A2 & 2 & rgb-only  & 0.3050 & 0.1050 & 0.0344 & 22.5311 \\
A3 & 3 & rgb-only  & 0.2556 & 0.0617 & 0.0174 & 26.3364 \\
A4 & 1 & Depth-3ch & 0.4398 & 0.1521 & 0.0693 & 11.6903 \\
A5 & 2 & Depth-3ch & 0.3980 & 0.1403 & 0.0574 & 14.1135 \\
A6 & 3 & Depth-3ch & 0.4200 & 0.1667 & 0.0714 & 12.8000 \\
A7 & 1 & Point cloud (XYZ)       & 0.4667 & 0.4532 & 0.2138 & 7.24 \\
A8 & 2 & Point cloud (XYZ)       & 0.4712 & 0.4602 & 0.2362 & 6.47 \\
A9 & 3 & Point cloud (XYZ)       & 0.4549 & 0.4674 & 0.2284 & 5.25 \\
\bottomrule
\end{tabular}
}
\end{table*}

\subsection{Main Results}
\label{sec:main_results}

We report results on touchable points and air points following the evaluation protocols in \cref{sec:bench}.
All air points metric results are reported conditional on relation correctness, since distance constraints are meaningful only when the intended geometric relation is satisfied.

\paragraph{Touchable Points.}
\label{sec:results_surface}
\cref{tab:surface_results} summarizes performance on RoboAfford-Eval for touchable points.
We report the official 2D accuracy across the three categories---\emph{Object Affordance Recognition} (OAR): identifying objects based on attributes such as category, color, size, and spatial relations, \emph{Object Affordance Prediction} (OAP): localizing functional parts of objects to support specific actions, such as the handle of a teapot for grasping, and \emph{Spatial Affordance Localization} (SAL): detecting vacant areas in the scene for object placement and robot navigation---as well as the overall average. Our experiments have demonstrated that our method achieves the sota performance among the current methods.
For methods that generate $(u,v,Z)$ outputs, we additionally report depth MAE (mm) against the monocular depth \emph{reference} used in our evaluation pipeline, with an inside/outside breakdown; methods that do not output depth are marked as ``$-$''.


\paragraph{Air points.}
\label{sec:results_free}
\cref{tab:free_overall} summarizes overall air points performance, while \cref{tab:free_bycat_micro} reports point-level micro results by category. \textbf{DirPt/MetPt} are point-level micro accuracies. 
\textbf{MetPt} and \textbf{MeanErr} are computed on direction-correct points in metric-offset queries (i.e., conditional on direction correctness). 
\textbf{FullPt} measures joint direction+metric success on metric-offset queries.
We report DirPt for relation correctness (direction-cone or between-cylinder), and for distance-constrained queries we additionally report MetPt@5cm, MeanErr (cm), and the joint success rate FullPt requiring both correct relation and distance satisfaction.
Following our protocol, MetPt@5cm and MeanErr are computed only on relation-correct points in distance-constrained queries.
For body-length constraints, the required offset is instantiated as a metric distance via the object-specific body-length scale derived from proxy geometry, enabling a unified 5\,cm tolerance across query types.
We include checkpoints fine-tuned for 1/2/3 epochs on the air points subset to study the effect of longer air points adaptation.
Results show consistent improvement with more training epochs.


\subsection{Ablations and Analysis}
\label{sec:ablation}

We ablate key design choices in our depth-aware formulation, including (A) whether depth is fed to the network, (B) alternative ways to feed depth, (C) the dual backbones design for depth tokens.
Unless otherwise specified, all variants follow the same training data and evaluation protocols as in \cref{sec:exp_setup} and \cref{sec:bench}.
\label{sec:ablation_settings}
\cref{tab:ablation_surface} and \cref{tab:ablation_free} summarize ablation results on touchable and air points targets, respectively.
For air points targets, we analyze two complementary axes: \textbf{relation correctness} and \textbf{distance fidelity conditioned on correct relation}.

\paragraph{\textbf{(A) Feed depth or not.}}
We remove all depth inputs to train and evaluate the model with RGB tokens and text tokens as input.
This variant tests whether depth cues are essential for 3D point prediction and geometric consistency.
The RGB-only variant drops on both touchable and air points benchmarks, indicating that explicit depth cues are important for executable 3D target prediction. (T1 \& T2 for touchable points in \cref{tab:ablation_surface} and A1 - A3 for air points in \cref{tab:ablation_free})

\paragraph{\textbf{(B) Alternative ways to feed depth.}}
We provide geometry as a lifted point cloud (from monocular depth) and convert it into a three-channel representation by directly quantizing $(x,y,z)$ values into three channels.
The resulting three-channel ``geometry map'' is fed to the model as an alternative geometry input, enabling a controlled comparison between \emph{depth-token fusion} and \emph{point-set-derived geometry input} under the same VLM interface. 
Compared with alternative geometry inputs, our depth-token fusion provides a simple and effective interface that improves relation correctness and metric precision. (T3-T6 \& A4-A9)

Notably, point-cloud input shows competitive distance accuracy on air points but weaker direction correctness, and performs substantially worse than the baseline on touchable points. Despite these limitations, point-cloud geometry offers complementary advantages worth further exploration in future work.

\paragraph{\textbf{(C) Shared vs.\ dual backbones.}}
We compare using a \emph{shared} backbone for both RGB and encoded depth images versus duplicating the backbone with separate parameters.
This ablation isolates whether a dedicated depth branch improves geometry token quality and downstream performance.
The dual-backbone design consistently outperforms the shared-backbone variant. (T1-T4)







\subsection{Qualitative Comparison}
\label{sec:qual}

We provide qualitative comparisons on RoboAfford-Eval to complement quantitative results.
\cref{fig:qual_surface} compares touchable-points predictions between our model and Qwen3-VL, where we visualize predicted points in the image (and depth-lifted 3D when applicable) to illustrate localization quality.
\cref{fig:qual_free} compares air points predictions, highlighting whether generated targets satisfy the intended spatial relation (direction/between) and distance constraints.

\section{Conclusion}
\label{sec:conclusion}

We introduced a minimalist execution interface for embodied tasks by formulating them as \emph{language-conditioned 3D target point prediction} with two complementary target types: \textbf{touchable} and \textbf{air points}.
Using open-source RGB images with monocular depth estimates, we constructed large-scale supervision and benchmarks under a unified $(u,v,Z)$ camera-centric representation.
Building on Qwen3-VL with an explicit depth-token stream, our model directly decodes structured point lists with the LM head and achieves consistent gains on both touchable and air points benchmarks under our evaluation protocols.

Current limitations include reliance on monocular depth estimates, which may degrade in textureless or reflective regions. Future work could extend embodied localization to dynamic scenes, integrate trajectory-level planning, and leverage world models to support richer spatial reasoning across embodied tasks.

\bibliographystyle{unsrtnat}
\bibliography{main}

\clearpage
\appendix

\begin{strip}
\centering
{\LARGE \bfseries SpatialPoint: Spatial-aware Point Prediction for Embodied Localization\par}
\vspace{0.5em}
{\large Supplementary Material\par}
\vspace{0.8em}
\end{strip}

\section{More Detailed Task Definitions and Evaluation Protocol}
\label{sec:supp_task_eval}

\subsection{Touchable and Air Points}
\label{sec:supp_touch_air}

Touchable points are surface-grounded 3D targets for direct physical interaction, while air points are free-space 3D targets specified by spatial language. Together, they provide a unified representation of action targets for both object contact and free-space interaction.

A valid air point must satisfy two constraints: it must match the instructed spatial relation and lie in unoccupied free space rather than on or inside occupied objects. This makes air-point prediction more challenging than direction-only grounding because the model must reason about both relative geometry and free-space validity.

\subsection{Unified Output Representation}
\label{sec:supp_uvz}

For both touchable and air points, the model predicts targets in the unified format $(u,v,Z)$, where $(u,v)$ denotes the image-plane location and $Z$ denotes metric depth. This representation keeps the prediction visually anchored while remaining directly executable in 3D.

Compared with full camera-frame 3D coordinates, $(u,v,Z)$ is more image-aligned: $(u,v)$ specifies the visually grounded location, while $Z$ provides the depth needed to recover the final 3D point.

\subsection{Detailed Geometric Evaluation for Air Points}
\label{sec:supp_air_eval}

As stated in the main text, all distance-related metrics for air points are evaluated \emph{only when the predicted point already satisfies the queried spatial relation}, since metric offsets are meaningful only for relation-valid predictions.

\paragraph{Object center.}
We use a unified object-center definition across all air-point categories. Each referenced object is lifted into 3D and represented by an object-level 3D proxy derived from its occupancy cue. The center of this proxy serves as the reference point for direction evaluation, between-object evaluation, and body-length-based metric computation.

\paragraph{Direction queries.}
For direction queries, we define a cone whose apex is the referenced object center and whose axis is the queried camera-frame direction. A prediction is considered relation-correct if it lies within $30^\circ$ of the cone axis, corresponding to a full angular aperture of $60^\circ$. This tolerance allows moderate ambiguity between neighboring directions and avoids an overly brittle criterion.

\paragraph{Between-object queries.}
For between-object queries, relation correctness is evaluated with respect to the line segment connecting the two referenced object centers. A prediction is considered relation-correct only if:
(i) its projection onto the segment lies between $10\%$ and $90\%$ of the segment length; and
(ii) its perpendicular distance to the segment is at most 10 cm.
This criterion encourages predictions that are genuinely between the two objects rather than merely close to the connecting line.

\paragraph{Body-length queries.}
For body-length queries, one body length is approximated as half of the diagonal length of the target object's 3D proxy box. A query such as \textit{two body lengths in front of the mug} is therefore converted into a target metric distance equal to twice this object-specific unit along the queried direction.

\paragraph{Occupancy validity.}
A valid air-point prediction must lie in free space. Predictions that fall inside occupied object volume are rejected, even if they satisfy the queried relation. This requires the model to reason about both spatial relations and basic scene occupancy.

\begin{figure*}[t]
    \centering
    \includegraphics[width=\textwidth]{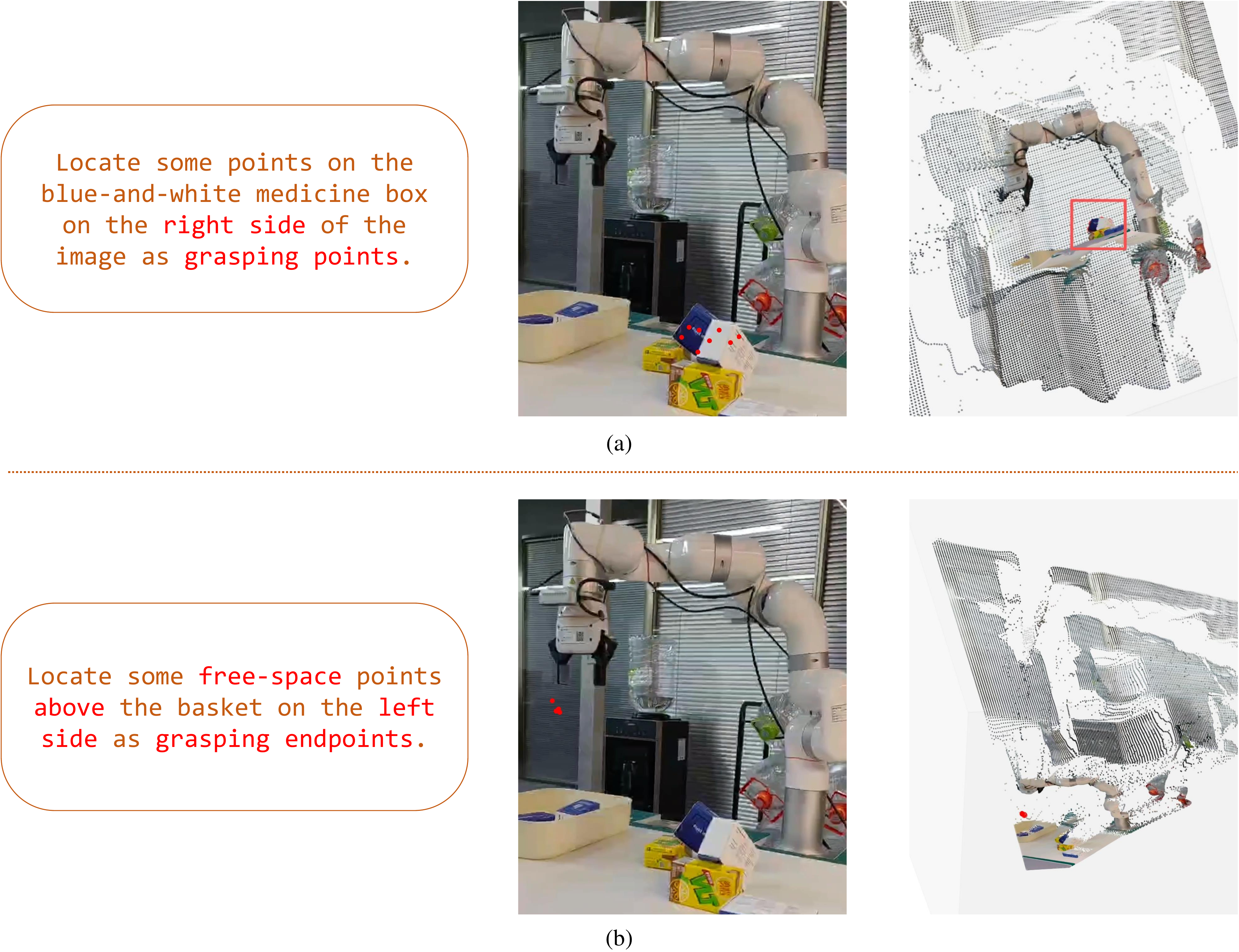}
    \caption{Zero-shot real-world results on robot pick-and-place.}
    \label{fig:supp_robot_3task}
\end{figure*}

\section{Zero-Shot Generalization to Real-World Robot Manipulation and Navigation}
\label{sec:supp_robot}

We provide additional real-world robot results to complement the main-text experiments on picking, placement, and navigation.
We emphasize that in both demonstrations, \textbf{SpatialPoint} operates without any fine-tuning on the target scene, highlighting its cross-scene generalizability.

\begin{figure*}[t]
    \centering
    \includegraphics[width=\textwidth]{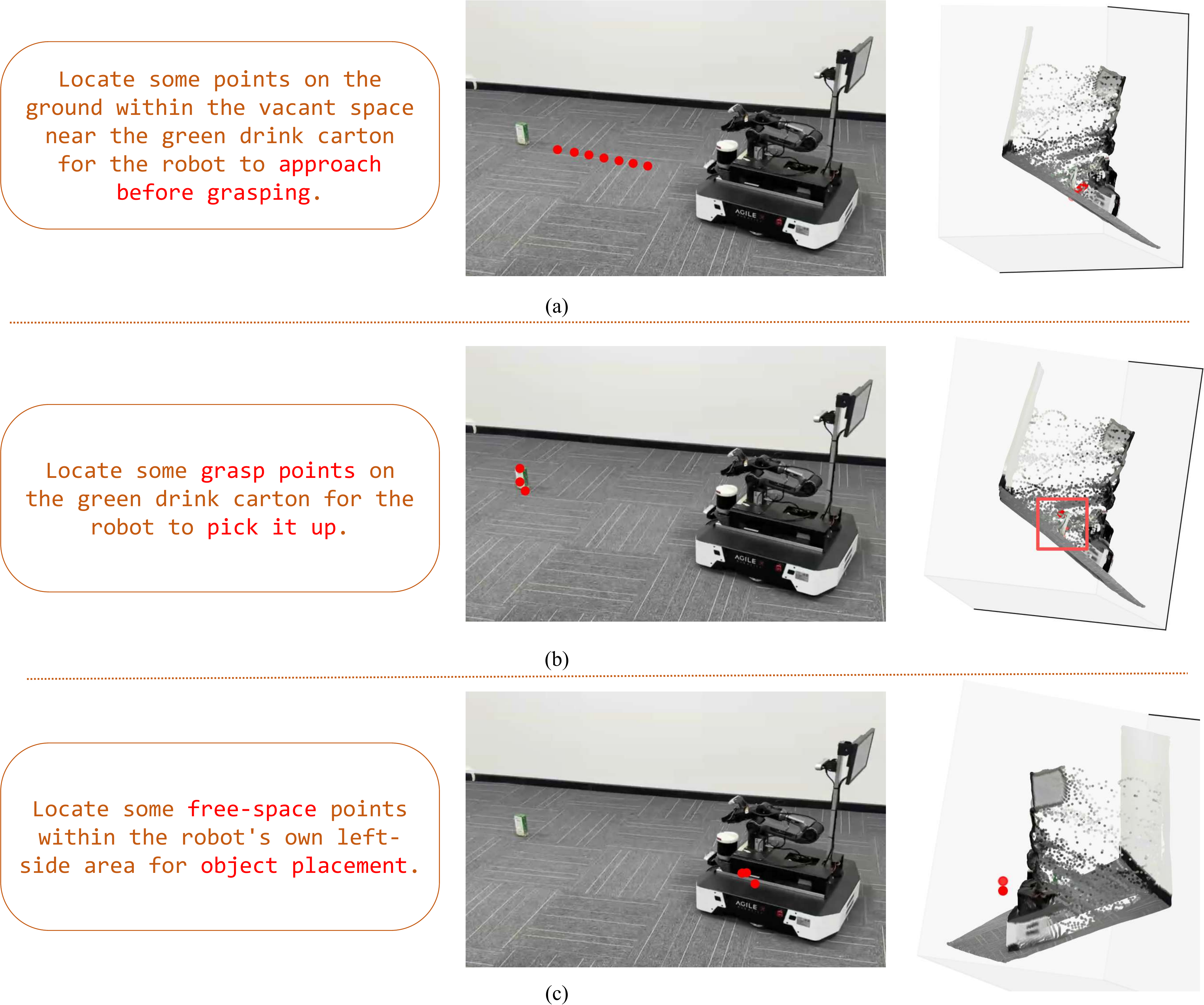}
    \caption{Zero-shot real-world robot results on mobile navigation and picking.}
    \label{fig:supp_robot_1scene_3task}
\end{figure*}
\noindent\textbf{Pick-and-Place.}
We deploy SpatialPoint on an uFactory xArm7 robotic arm \cite{ufactory_xarm7} in a tabletop setting with multiple boxes.
SpatialPoint serves two roles: (1) predicting the grasping contact point on the target box via natural language instruction (\textit{touchable point}, \cref{fig:supp_robot_3task}(a)), and (2) predicting the target placement position for the grasped object via natural language instruction (\textit{air point}, \cref{fig:supp_robot_3task}(b)).
We refer the reader to our project page, where the complete manipulation sequence video (\texttt{1-robotarm-pick-place.mp4}) is provided; in this video, the robot arm localizes the target box and places it at the specified destination.

\noindent\textbf{Mobile Manipulation.}
We deploy SpatialPoint on a mobile robot built upon AGILEX Tracer 2.0 \cite{agilex_tracer2} and AGILEX Piper \cite{agilex_piper}.
SpatialPoint supports three sequential subtasks within a single scene: (1) predicting the approach position near the target object to enable grasping (\textit{touchable point}, \cref{fig:supp_robot_1scene_3task}(a)), (2) localizing the target object for interaction (\textit{touchable point}, \cref{fig:supp_robot_1scene_3task}(b)), and (3) predicting the placement position for the grasped object (\textit{air point}, \cref{fig:supp_robot_1scene_3task}(c)).
We refer the reader to our project page, where the complete process video (\texttt{2-mobile-navigation.mp4}) is provided; in this video, the mobile robot navigates to the target location and retrieves the bottle.


\begin{figure*}[t]
    \centering
    \includegraphics[width=\textwidth]{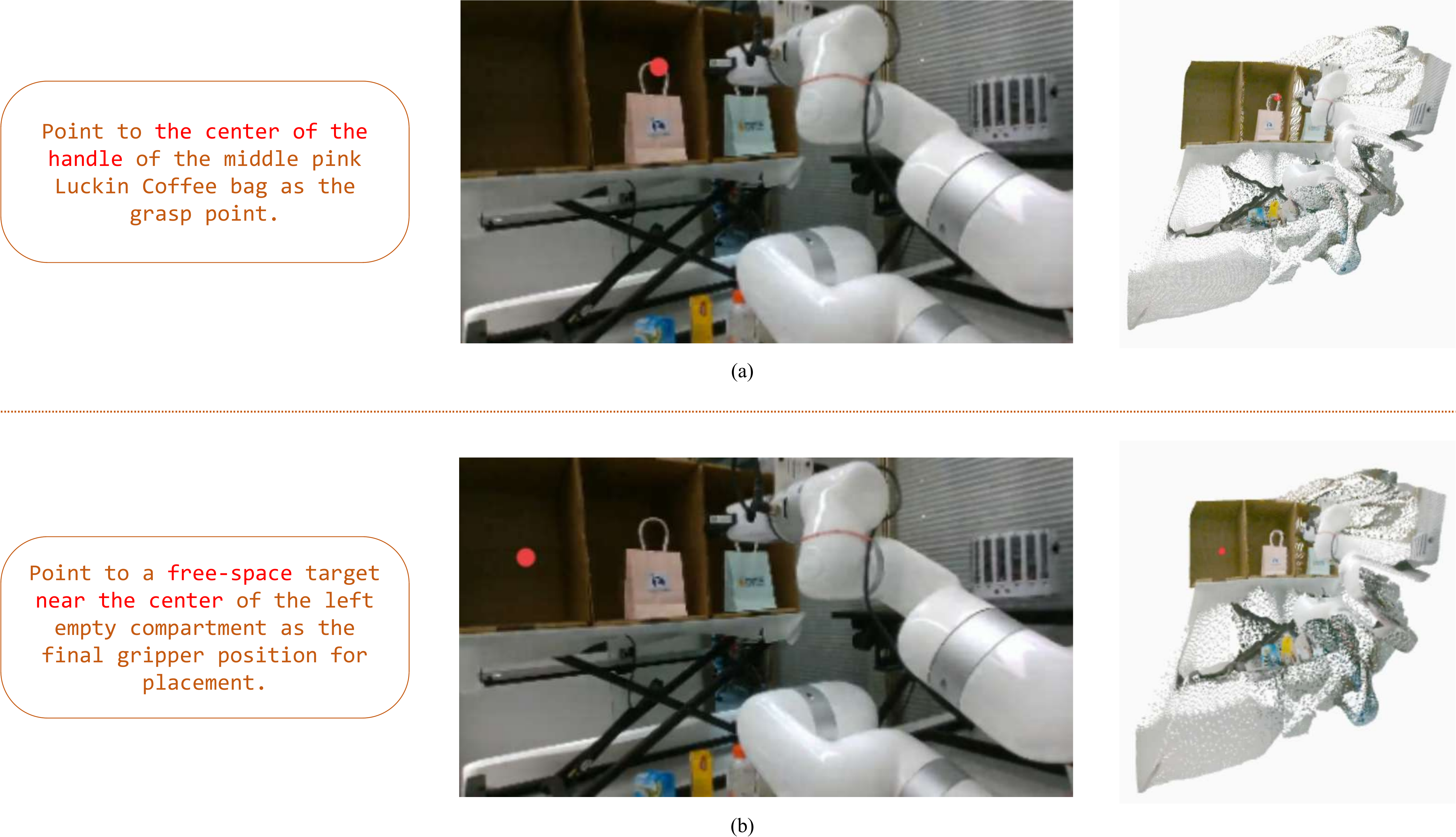}
    \caption{Zero-shot real-world robot results on long-horizon task.}
    \label{fig:supp_robot_long}
\end{figure*}

\noindent\textbf{Long-Horizon Task: Package Reordering.}
We apply SpatialPoint to a long-horizon manipulation task involving the reordering of multiple packages within a multi-compartment open shelf.
The packages are placed in individual compartments, and a uFactory xArm7 robotic arm~\cite{ufactory_xarm7} is instructed to reorder them according to human voice or text commands.
SpatialPoint supports two sequential subtasks: (1) predicting the grasping point on the target package (\textit{touchable point}, \cref{fig:supp_robot_long}(a)), and (2) predicting the placement destination within the specified compartment following the human instruction (\textit{air point}, \cref{fig:supp_robot_long}(b)).
We refer the reader to our project page, where the complete manipulation sequence video (\texttt{3-longhorizon-pack-reranking.mp4}) is provided.

\section{Additional Experimental Results}
\label{sec:supp_additional_results}

This section presents additional qualitative comparisons and ablation results. 
We extend the qualitative comparison by adding RoboBrain2.5~\cite{tan2026RoboBrain} as an extra baseline alongside Qwen3-VL-4B~\cite{bai2025qwen3} and our model, and we further report a supplementary air-point ablation under SpatialBot-style depth encoding.

\subsection{More Qualitative Comparisons}
\label{sec:supp_more_qual}

Compared with the main text, we additionally include RoboBrain2.5~\cite{tan2026RoboBrain} as a baseline and present three-model qualitative comparisons under identical instructions.

\begin{figure*}[t]
    \centering
    \includegraphics[width=\textwidth]{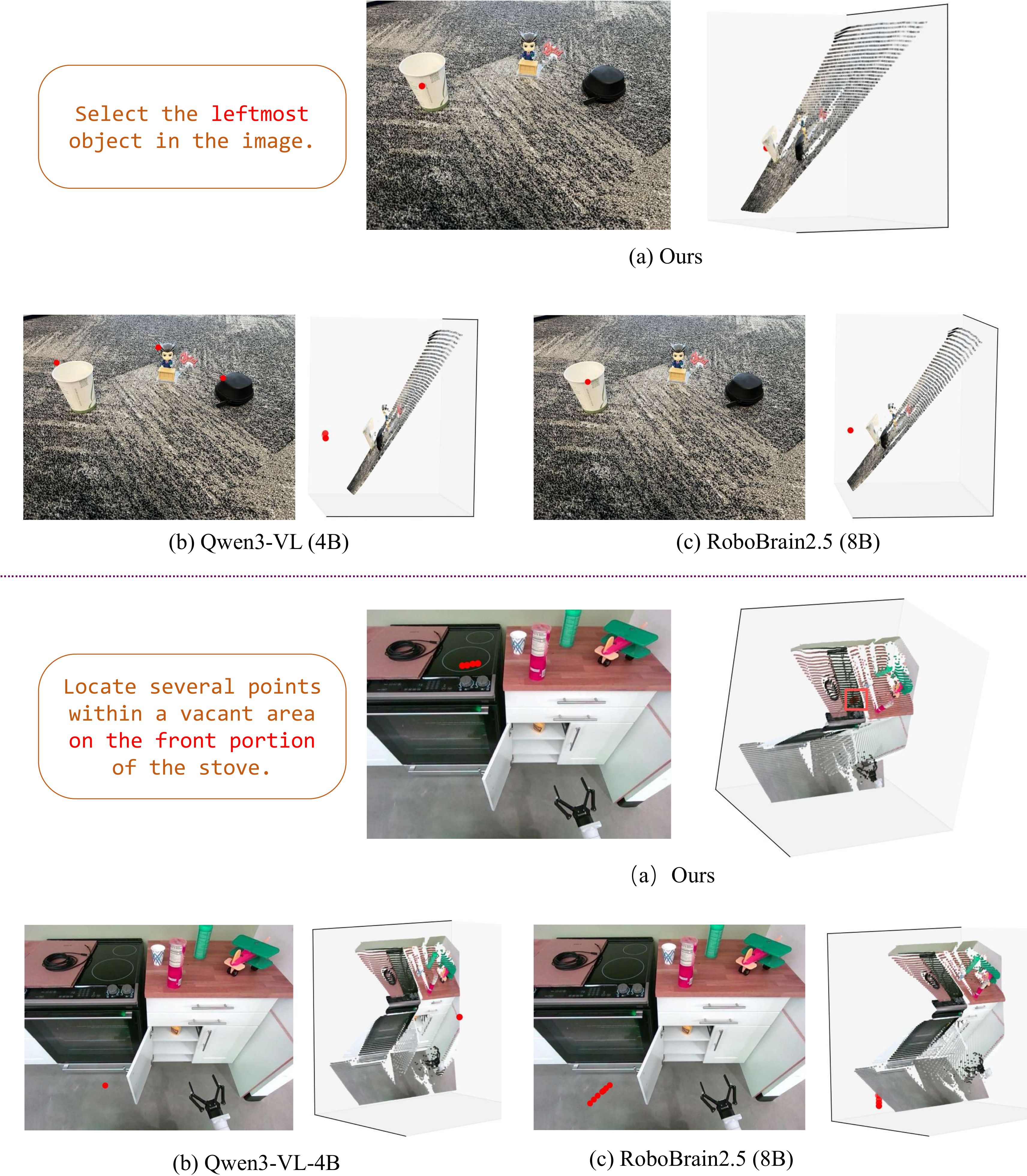}
    \caption{Additional qualitative comparisons on touchable points prediction. We compare 3 models under the same instructions.}
    \label{fig:supp_touchable_more}
\end{figure*}

\begin{figure*}[t]
    \centering
    \includegraphics[width=\textwidth]{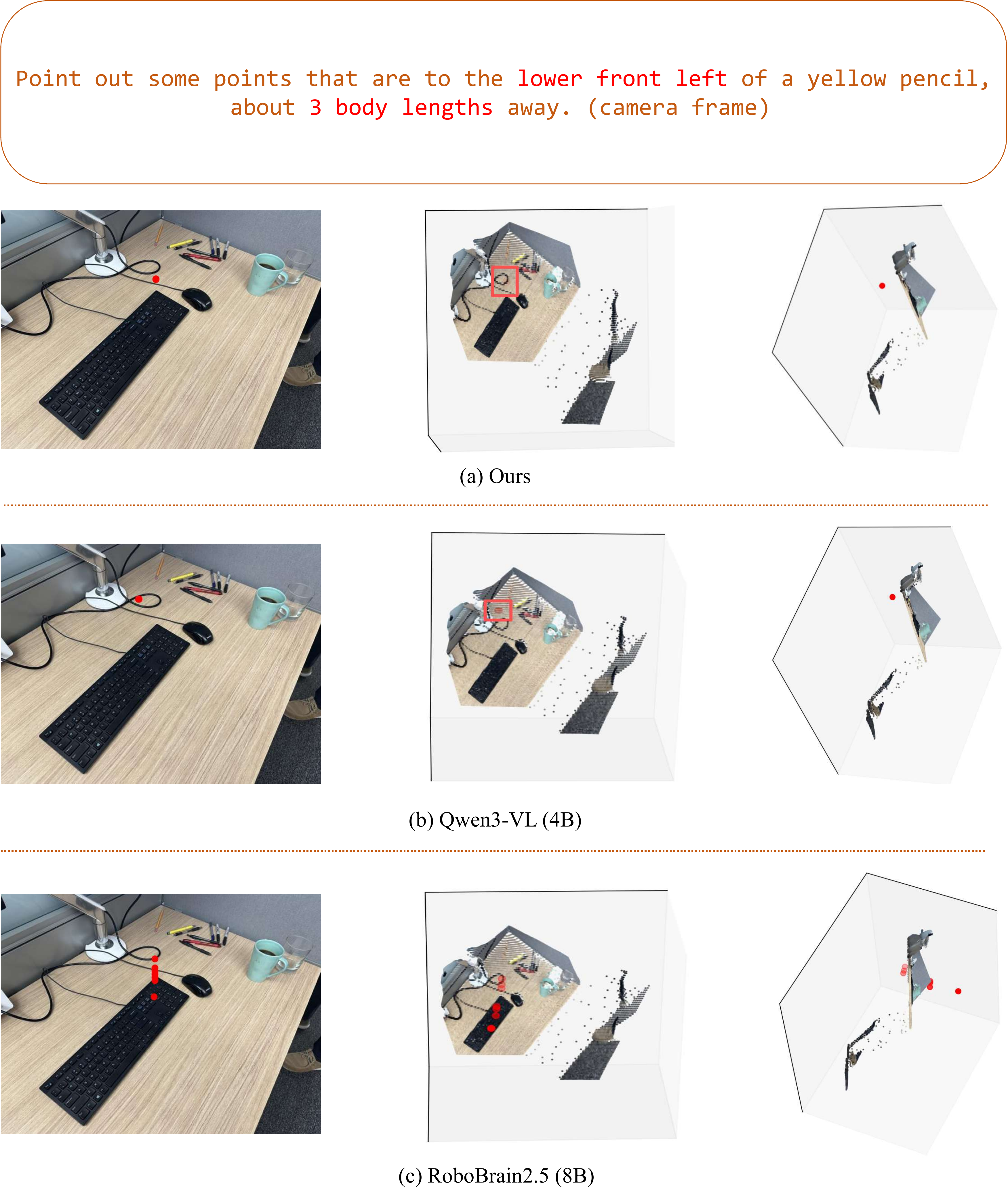}
    \caption{Additional qualitative comparisons on air points prediction. We compare 3 models under the same instructions.}
    \label{fig:supp_air_more}
\end{figure*}

\cref{fig:supp_touchable_more} shows additional touchable-point examples, while \cref{fig:supp_air_more} shows additional air-point examples.

\subsection{Supplementary Ablations on Air Points}
\label{sec:supp_air_ablation_spatialbot}

Owing to space limitations, the main text reports only the default air-point ablation setting. Here we include additional ablation results under SpatialBot-style depth encoding to separately show the effects of the dual-backbone design and special depth tokens. All variants use the same SpatialBot-style depth representation and are trained for 3 epochs.

\cref{tab:supp_ablation_air_spatialbot} summarizes the results. Compared with the condensed presentation in the main text, these additional results provide a more explicit breakdown of the two architectural choices, since the default setting there already includes both components.

\begin{table*}[t]
\centering
\caption{Supplementary air-point ablations under SpatialBot-style depth encoding.}
\label{tab:supp_ablation_air_spatialbot}
\small
\setlength{\tabcolsep}{5.5pt}
\renewcommand{\arraystretch}{1.08}
\resizebox{\textwidth}{!}{%
\begin{tabular}{lccccccc}
\toprule
\textbf{Variant} & \textbf{Epoch} & \textbf{Dual Backbone} & \textbf{Special Tokens} & \textbf{DirPt$\uparrow$} & \textbf{MetPt@5cm$\uparrow$} & \textbf{FullPt$\uparrow$} & \textbf{MeanErr (cm)$\downarrow$} \\
\midrule
Ours1 & 1 & $\checkmark$ & $\checkmark$ & 0.4886 & 0.2587 & 0.1300 & 8.5008  \\
Ours2 & 2 & $\checkmark$ & $\checkmark$ & \textbf{0.5088} & \underline{0.2907} & \underline{0.1456} & \underline{7.3034}  \\
Ours3 & 3 & $\checkmark$ & $\checkmark$ & \underline{0.5071} & \textbf{0.3347} & \textbf{0.1641} & \textbf{6.8084}  \\
\midrule
A10 & 1 & $\times$ & $\times$     & 0.4088 & 0.1885 & 0.0795 & 11.2143 \\
A11 & 2 & $\times$ & $\times$ & 0.4413 & 0.1916 & 0.0846 & 11.2934\\
A12 & 3 & $\times$ & $\times$  & 0.4403 & 0.2130 & 0.0937 & 10.2762\\
A13 & 1 & $\times$ & $\checkmark$     & 0.4315 & 0.1913 & 0.0854 & 11.5730\\
A14 & 2 & $\times$ & $\checkmark$ & 0.4576 & 0.2453 & 0.1134 & 10.9660\\
A15 & 3 & $\times$ & $\checkmark$  & 0.4650 & 0.2482 & 0.1164 & 9.7431\\
A16 & 1 & $\checkmark$ & $\times$     & 0.4252 & 0.1911 & 0.0827 & 10.5795\\
A17 & 2 & $\checkmark$ & $\times$ & 0.4543 & 0.2299 & 0.1043 & 10.8764\\
A18 & 3 & $\checkmark$ & $\times$  & 0.4609 & 0.2454 & 0.1133 & 10.0685\\
\bottomrule
\end{tabular}
}
\end{table*}

Overall, the trends are consistent with those reported in the main paper and further clarify the contribution of the dual-backbone design and special depth tokens under SpatialBot-style depth encoding.

\section{Qualitative Analysis of Failure Cases}
\label{sec:supp_failure}

\subsection{Failure Analysis on Touchable Points}
\label{sec:supp_failure_touchable}

We observe three common failure patterns in touchable-point prediction, corresponding to progressively more severe errors in localization, constraint satisfaction, and target grounding.

\begin{figure*}[t]
    \centering
    \includegraphics[width=\textwidth]{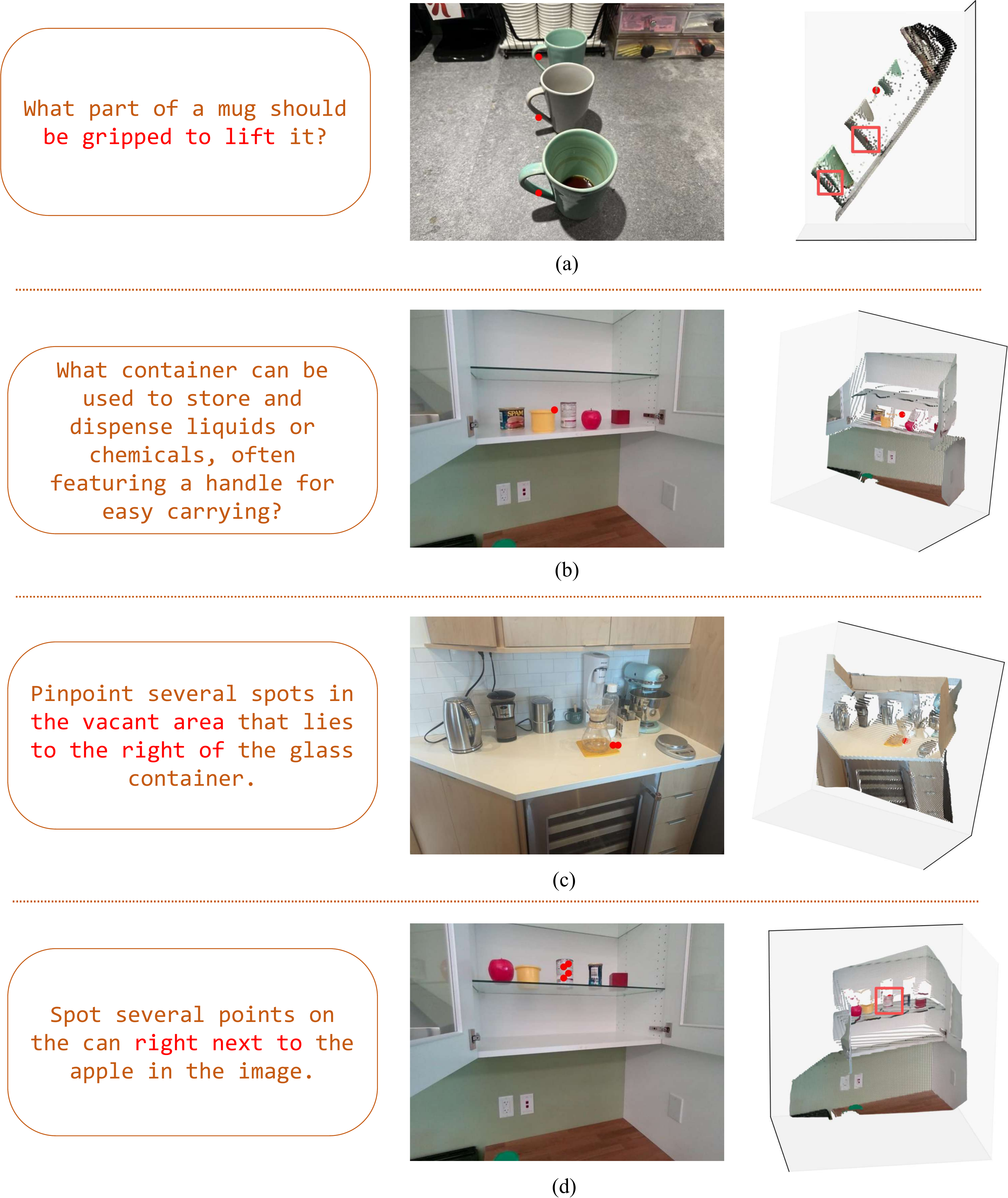}
    \caption{Failure cases on touchable-point prediction. (a) and (b) Fine-grained localization errors. (c) Fine-grained constraint error. (d) Relational grounding error.}
    \label{fig:supp_failure_touchable}
\end{figure*}

\paragraph{Fine-grained localization error.}
In some cases, the model identifies the correct target region but predicts a slightly shifted point, as illustrated in~\cref{fig:supp_failure_touchable}(a) and (b). Such errors are common near object boundaries or thin structures, where even a small image-plane deviation can lead to a noticeable 2D mismatch and a larger 3D error after depth lookup. This also helps explain why in-mask depth errors are typically smaller than out-of-mask ones: small boundary shifts can already destabilize both image-plane alignment and the associated depth value.

\paragraph{Fine-grained constraint error.}
A second failure mode occurs when the model captures the coarse spatial relation but fails to satisfy the finer constraint in the instruction, as shown in~\cref{fig:supp_failure_touchable}(c). For example, when the instruction refers to a vacant region on one side of an object, the prediction may fall on the correct side but not in a truly vacant or interaction-valid area. These cases suggest that touchable-point prediction requires not only coarse directional understanding, but also finer discrimination of locally valid target regions.

\paragraph{Relational grounding error.}
A third and more severe failure mode arises when the model misidentifies the referred target or misgrounds the underlying spatial relation, as shown in~\cref{fig:supp_failure_touchable}(d). In such cases, the prediction is anchored to the wrong object or relational target rather than being a small local deviation.

\subsection{Failure Analysis on Air Points}
\label{sec:supp_failure_air}

Air-point prediction is more challenging than touchable-point prediction because the target lies in free space and must satisfy both spatial and occupancy constraints. We therefore focus on representative relation-level failures and leave metric errors aside, as they are harder to interpret in isolation.

\begin{figure*}[t]
    \centering
    \includegraphics[width=\textwidth]{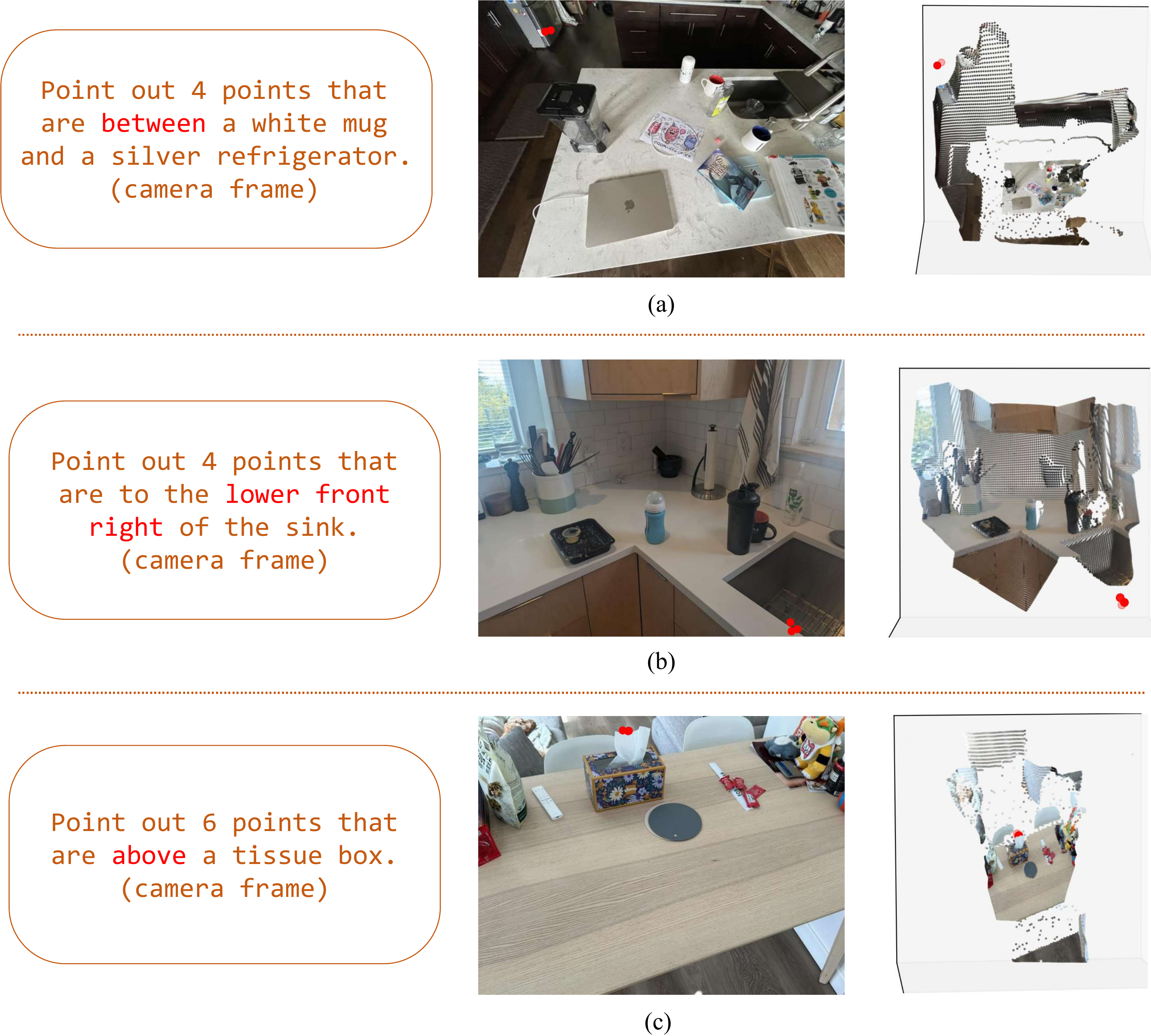}
    \caption{Failure cases on air-point prediction. (a) Relational direction misinterpretation. (b) Weak image support for free-space targets. (c) Occupancy-awareness failure.}
    \label{fig:supp_failure_air}
\end{figure*}

\paragraph{Relational direction misinterpretation.}
In this case, the model recognizes the referenced object but fails to place the target in the instructed relational direction, as shown in~\cref{fig:supp_failure_air}(a). This indicates that air-point prediction requires not only identifying the correct reference object, but also accurately grounding directional relations in 3D.

\paragraph{Weak image support for free-space targets.}
In this case, the queried free-space target is geometrically well defined in camera-frame 3D space but only weakly supported by the visible 2D image, especially near image boundaries, as shown in~\cref{fig:supp_failure_air}(b). This highlights a key difficulty of air-point prediction: a target may be clear in 3D yet hard to infer from the image plane alone.

\paragraph{Occupancy-awareness failure.}
In this case, the model captures the coarse relational cue but predicts a point that still falls inside or too close to occupied object space, as shown in~\cref{fig:supp_failure_air}(c). This indicates that successful air-point prediction requires not only correct directional reasoning, but also stronger awareness of local 3D occupancy and free-space feasibility.

\end{document}